\pdfoutput=1

\documentclass[11pt]{article}

\usepackage[final]{acl}

\usepackage{times}
\usepackage{latexsym}

\usepackage[T1]{fontenc}

\usepackage[utf8]{inputenc}

\usepackage{microtype}
\usepackage{algpseudocode}
\usepackage{algorithm}

\usepackage{xcolor} 
\usepackage{colortbl}
\usepackage{inconsolata}
\usepackage{amsmath}
\usepackage{multirow}

\DeclareMathOperator*{\argmin}{arg\,min}
\DeclareMathOperator*{\argmax}{arg\,max}
\usepackage{graphicx}
\usepackage{subcaption}

\usepackage{tabularx}
\usepackage{booktabs} 
\usepackage[skip=3pt]{caption} 
\usepackage{array} 
\title{Mitigating Confounding in Speech-Based Dementia Detection through Weight Masking}

\author{
  Zhecheng Sheng\textsuperscript{†} \quad Xiruo Ding\textsuperscript{*} \quad Brian Hur\textsuperscript{*} \quad Changye Li\textsuperscript{*} \\
    {\bf Trevor Cohen}\textsuperscript{*} \quad {\bf Serguei Pakhomov}\textsuperscript{†} \\
  \textsuperscript{†}University of Minnesota \quad \textsuperscript{*}University of Washington \\
  \textsuperscript{†}\texttt{\{sheng136, pakh0002\}@umn.com}
}

\begin{document}
\maketitle
\begin{abstract}
Deep transformer models have been used to detect linguistic anomalies in patient transcripts for early Alzheimer's disease (AD) screening. While pre-trained neural language models (LMs) fine-tuned on AD transcripts perform well, little research has explored the effects of the gender of the speakers represented by these transcripts. This work addresses gender confounding in dementia detection and proposes two methods: the \textit{Extended Confounding Filter} and the \textit{Dual Filter}, which isolate and ablate weights associated with gender. We evaluate these methods on dementia datasets with first-person narratives from patients with cognitive impairment and healthy controls. Our results show transformer models tend to overfit to training data distributions. Disrupting gender-related weights results in a deconfounded dementia classifier, with the trade-off of slightly reduced dementia detection performance.
\end{abstract}

\section{Introduction}
Transformer-based models \citep{vaswani2017attention} have excelled in language and vision tasks, particularly bidirectional encoder models like BERT \citep{devlin-etal-2019-bert} and its variants \citep{liu2019roberta, sanh2020distilbert, lee2020biobert, qian2022perturbation}, which enhance classification through rich text representations.
As these models gain traction in clinical tasks like dementia detection, it is crucial to ensure fairness in their predictions, given the high stakes of clinical decision-making. However, most models are optimized for task-specific accuracy without accounting for biases present in fine-tuned datasets \citep{baldini2022fairness, bolukbasi2016man, hutchinson2020social, webster2021measuring, manela2021stereotype}, leading to spurious correlations. 

Efforts to mitigate these biases have focused on two main approaches. One involves task-agnostic methods that enforce fair representation learning \citep{kaneko2021debiasing, cheng2021fairfil, guo2022autodebias}, while the other targets bias reduction in specific tasks using annotated data \citep{shen2021contrastive, ravfogel2022linear, gira-etal-2022-debiasing, zhu2023debiased}.
A particularly challenging form of bias is confounding bias \citep{landeiro2018conf}, which arises when extraneous factors distort the relationship between the input language and the diagnostic outcomes. In spoken language-based dementia assessment, the existence of confounders can influence both linguistic patterns and disease prevalence, leading models to learn unintended associations. 
Despite growing awareness of bias in machine learning, confounding bias in \textit{low-resource} domains like healthcare remains underexplored, where imbalanced datasets exacerbate the problem.

In this study, we investigate gender confounding bias in dementia detection using speech-based datasets. Prior research in Alzheimer’s disease has identified sex as a prominent risk factor, with significant differences in dementia incidence observed between males and females \citep{beam2018differences, Podcasy31122016}. While such pathological disparities are clinically meaningful, machine learning models that predict dementia from speech should make predictions independently of gender differences. A biased model may rely on gender-specific language cues rather than clinically relevant markers of cognitive decline, potentially leading to misdiagnoses and unequal performance across demographic groups.

To address this, we introduce two novel bias mitigation techniques inspired by the Confounding Filter \citep{wang2019removing}: Extended Confounding Filter (ECF) and Dual Filter (DF). We evaluate these methods on two dementia speech datasets\footnote{Due to limited public datasets for Alzheimer's disease (AD) classification, we also considered \texttt{ADReSS} \citep{luz2020alzheimersdementiarecognitionspontaneous} but excluded it due to its small size compared to the other two.} widely used in cognitive linguistic research \citep{li-etal-2022-gpt, farzana-parde-2023-towards}. Our main contributions in this paper are as follows \footnote{Our code is available at \url{https://github.com/LinguisticAnomalies/DualFilter.git}.}:
\begin{enumerate}
\item We identified under-explored gender confounding bias in speech datasets for dementia. 
\item We extended the Confounding Filter method to the Transformers architecture and demonstrated improvements in downstream task performance.
\item We introduced the Dual Filter as a simple yet effective weight masking algorithm  that identifies and ablates parameters associated with the confounding bias in the entire model's network (vs. individual layers).
\item We showed that both proposed methods effectively reduce the False Positive Rate (FPR) and Statistical Parity (SP) gap between genders while maintaining relatively strong model performance under various distribution shifts.
\end{enumerate}

\section{Related Work}
In recent years, transformer-based models have demonstrated promising performance in dementia detection using patient speech data \citep{hernandez2018computer, cohen2020tale, LuzHaider20ADReSS, guo2021cookie, li-etal-2022-gpt}. However, these models are susceptible to inductive bias due to the small size of publicly available datasets utilized in most studies. A key concern is that these models may learn gender-specific language patterns from male and female participants performing the same task, and subsequently use these differences to make dementia predictions, regardless of the participants' true cognitive status.

The methods we propose involves isolating and removing the influence of model weights associated with a confounding variable. As such, our work relates to prior efforts aimed at regularizing information encoded within transformer networks. One line of research explores weight isolation through disentangled learning \citep{zhang-etal-2021-disentangling, colombo-etal-2021-novel}, which require specialized loss functions to minimize information overlap between targets and sensitive attributes. Adapter sub-networks, on the other hand, regulate and control access to information from protected features \citep{hauzenberger-etal-2023-modular, masoudian-etal-2024-effective}. Without introducing additional training objectives, \citet{liu2024the} propose a gradient integration method to identify neurons responsible for disparities in output logit distributions among demographic groups. \citet{lee2019snip} and \citet{sun2024simple} develop weight importance ranking algorithms to locate redundant weights for network pruning. These algorithms track neuron activations or loss outputs by masking certain weights within a layer and assigning importance scores based on a calibrated dataset.

While most of these methods focus on pretrained models—some exceptions serve as baselines in our experiments—our work specifically addresses bias learned during fine-tuning for a given task. Compared to prior approaches, our method for identifying influential weights integrates seamlessly into standard training, requiring no additional components or objective formulations, yet achieving significant bias mitigation across various distribution shifts in the data.

\section{Methods}

\begin{figure*}[ht!]
    \centering
    \begin{subfigure}[b]{0.50\textwidth}
        \centering
        \includegraphics[width=\textwidth]{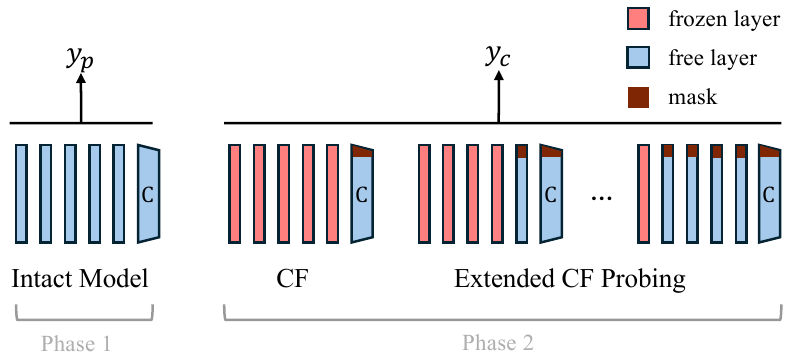}
        \caption{}
        \label{fig:exconf}
    \end{subfigure}
    \hfill
    \begin{subfigure}[b]{0.45\textwidth}
        \centering
        \includegraphics[width=\textwidth]{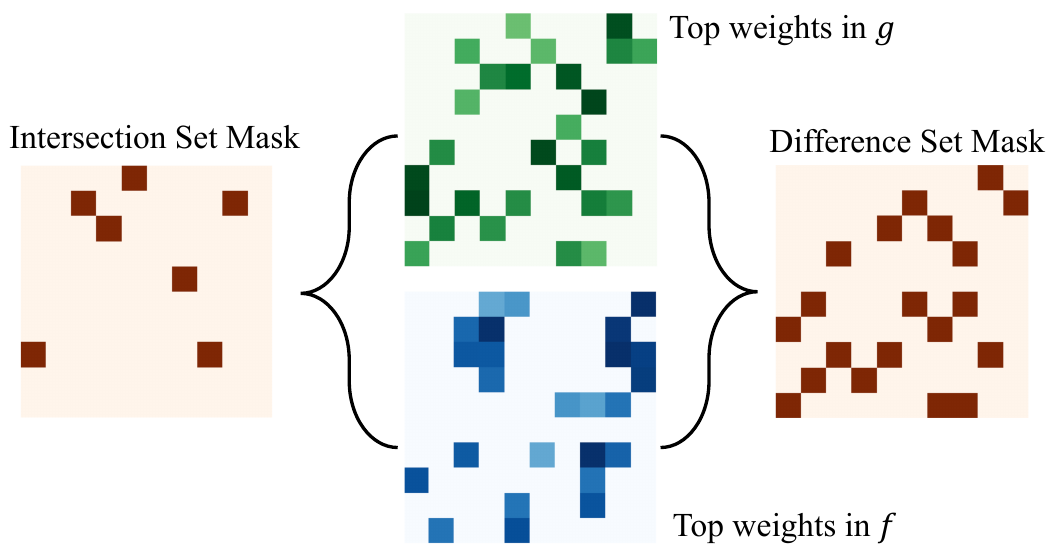}
        \caption{}
        \label{fig:dfalg}
    \end{subfigure}
    \caption{\textbf{(a)} Illustration of the Extended Confounding Filter (\textbf{ECF}) Probing framework for weights identification. \textbf{(b)} Illustration of the Dual Filter (\textbf{DF}) procedure to find weights to mask. }
    \vspace{-4mm}
    \label{fig:fig1}
\end{figure*}

\subsection{Confounding Filter}\label{sec:cf}
Deep learning models often recognize false signals from confounding factors, leading to sub-optimal performance in many real-world cases \citep{szegedy2013intriguing, nguyen2015deep, wang2017origin, wang2017select}. To address this issue, the Confounding Filter \citep{wang2019removing} was proposed to address confounding biases in models trained on electroencephalogram and medical imaging data. In this approach, a deep learning model is denoted as having two components: $g(\cdot ; \theta)$, a representation learning network, and $f(\cdot ; \phi)$, a classification network. The algorithm first optimizes the entire network by solving the following objective:

\begin{equation*}
    \hat{\theta}, \hat{\phi} = \underset{\theta, \phi}{\argmin}\:\mathcal{L}(y, f(g(X); \theta);\phi),
\end{equation*}

\noindent where $\mathcal{L}$ denotes the loss function to be minimized. 

In the second phase, assuming we have access to the confounder label $m$ in the dataset, the algorithm localizes weights that are reactive to the confounding variable. This is achieved through tuning $f(\cdot;\phi)$ towards $m$ while keeping $g(\cdot;\theta)$ fixed. During the second phase, updates in $\hat{\phi}$ are tracked and normalized after each batch. The sum of normalized updates is denoted as $\pi = \frac{1}{b}\sum_{i=1}^{b}\lvert\Delta\phi_i\rvert$ where $b$ is the number of total batches in the second phase of training. The importance of each element in $\pi$ is determined by their magnitude. A threshold function is then employed to get the mask:
\[
M_i =
\begin{cases} 
0 & \text{if } \pi_i > \tau \\
1 & \text{otherwise}
\end{cases}
\]

Here, $\tau$ is the $k^{th}$ percentile in $\pi$, where $k$ is a hyperparameter. The element-wise product $\hat{\phi^{'}} = \hat{\phi} \otimes M$ results in the confounder-mitigated network $f(g(X); \hat{\theta});\hat{\phi^{'}})$.

\subsection{Extended Confounding Filter}
While the original Confounding Filter algorithm has shown improvements over the baseline in some neural network architectures \citep{wang2019removing}, its adaptation to transformer networks remains unexplored. Transformer-based language models (LMs) learn to generate distributional semantic representations \citep{vaswani2017attention} through the attention mechanism and positional encoding. By fine-tuning a pretrained LM, semantic information pertinent to a task of interest is dynamically stored across the transformer network layers.

Our hypothesis is that fixing $g(\cdot;\theta)$ when training for the confounder variable may not effectively capture the most confounder-associated weights within the transformer network. To test this hypothesis, we sequentially unfroze each layer in the transformer network, starting from the classification head down to the embedding layer and observed its impact on the prediction. This is different from the original Confounder Filter method, where only the classification head is trainable in the encoder model. We refer it as Extended Confounding Filter (ECF) in the paper.

\begin{figure}[htbp]
        \centering
        \includegraphics[width=0.45\textwidth]{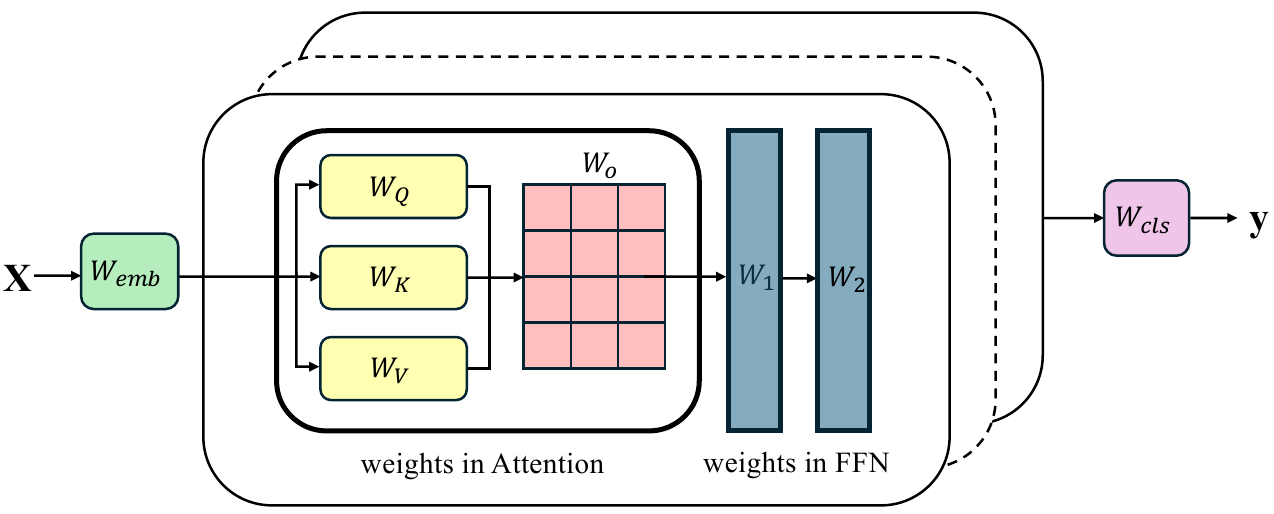}
        \caption{Tracked weights in the transformer network}
        \label{fig:trackedweights}
\end{figure}

The illustration of how ECF works is shown in Figure~\ref{fig:trackedweights}. Matrices $W_Q, W_K, W_V, W_O, W_1, W_2$ are tracked in a single transformer block, while $W_{emb}$ and $W_{cls}$ represent the token embedding matrix and classification weight matrix in a sequence classification model, respectively. Similarly to the Confounding Filter, we start by training a classification model towards the primary label $Y_p$ (Phase 1) and then continue training the model towards classifying the confounder label $Y_c$ with layers sequentially unfreeze (Phase 2).

\vspace{-2mm}
By sequentially unfreezing different numbers of layers, we allow varying amounts of the model's parameter spaces to react to the information introduced during Phase 2 (Figure~\ref{fig:exconf}). The sequential probing scheme follows the idea of the Confounding Filter but offers greater flexibility, as it allows partitioning of the classification network $f(x)$ and representation learning network $g(x)$ at different points.
The change in model parameters $\Delta\phi_i$ is normalized within the matrix and recorded after each training batch. Following the Confounding Filter methodology, we restrict $\Delta\phi_i$ to each $W$ in this probing procedure, and the threshold $\tau$ is calculated for each individual weight matrix. The probing step size is by layer. Masking matrices, derived from the threshold function, are applied to the tracked weight matrix from Phase 1 fine-tuning. 

\subsection{Dual Filter}
We further relax the restriction in Phase 2 training of the ECF method, which mandates local masking and overlooks the dynamic interactions within the LM during fine-tuning. To address this, we propose Dual Filter, a method that tracks weight changes from two separate models initialized from the same checkpoint—one optimized for the primary outcome (\textit{dementia}) and the other for the confounder (\textit{gender}). After obtaining change matrices $\pi$ from both models, we utilize set operations to isolate weights that are most reactive to the confounder label during finetuning. Specifically, we choose the top $k\%$ most changed weights from the primary model $f$ and the confounder model $g$, and take the intersection or the difference from these two weight sets to generate the mask matrices (Figure~\ref{fig:dfalg}). One could strategically apply the intersection set mask ($M_I$), the difference set mask ($M_D$), or the joint set ($M_I \cup M_D$) of both masks, which is equivalent to selecting the top k\% most changed weights from the confounder model, depending on the dataset and task. This flexibility allows for precise bias mitigation, ensuring optimal trade-offs between fairness and performance tailored to specific applications. We formally describe the proposed method in Algorithm~\ref{alg:dual}.

\begin{algorithm}
\caption{Dual Filter for weights masking}\label{alg:dual}
\begin{algorithmic}[1]
\Require pretrained LM: $f_0(x)$, $g_0(x)$; dataset: $\mathcal{D}(x, y_p, y_c)$; threshold: $k$
\Ensure Confounder-adjusted model $f(x;\theta^{'})$

\State Train $f_0(x;\theta) \mapsto y_p$, obtain weights change $\Delta_p$ and fine-tuned model $f(x;\hat\theta)$. 
\State Train $g_0(x;\phi) \mapsto y_c$, obtain weights change $\Delta_c$ and fine-tuned model $g(x;\hat\phi)$.
\State  \[
\Delta_{p, k} = \argmax_{p \subseteq \Delta_p, |p| = k} \sum_{p_i \in \Delta_p} p_i
\]
\[
\Delta_{c, k} = \argmax_{c \subseteq \Delta_c, |c| = k} \sum_{c_i \in \Delta_c} c_i
\]
\State $M_I \gets \Delta_{p, k} \cap \Delta_{c, k}$, $M_D \gets \Delta_{c, k} \setminus \Delta_{p, k} $
\State Pick mask $M\in \{M_I, M_D, M_I \cup M_D\}$
\State \[
\theta^{'} \gets \hat\theta_i = 0 \quad \forall i \in M
\]
\end{algorithmic}
\end{algorithm}
\subsection{Other Baselines}
We include two recent baseline adapter models, \textsc{ConGater} \citep{masoudian-etal-2024-effective} and \textsc{ModDiffy} \citep{hauzenberger-etal-2023-modular} using their publicly available code\footnote{\url{https://github.com/ShawMask/DebiasingConGater}}$^{,}$\footnote{\url{https://github.com/CPJKU/ModularizedDebiasing}}, both of which address biases learned during fine-tuning. These methods append additional bias-mitigate modules into a network and update their weights through a joint loss function. We run these two methods using their default configurations with a range of hyperparameters (See Table~\ref{tab:baseline_hyp}) to compare their fairness-performance trade-offs with the proposed methods.

\section{Evaluations} \label{sec:evalmethod}
\paragraph{Confounding Shift} One fundamental assumption in machine learning is that the test and training datasets are from the same distribution. However this assumption is often violated in real world applications resulting in distribution shifts. One specific form of distribution shift is sub-population shift \citep{cao2019learning, cai2021theory}, where the training distribution differs from the deployment distribution. A model trained on such shifted data tends to learn spurious correlations with the majority class, resulting in poor performance when applied to data with a class distribution different from that of the training set \citep{yang2023change}. 

While the sub-population shifts are determined by the product of group attributes and the label, and the group attributes are not independent of the label, it is a special type of dataset shift referred to as \textit{Confounding Shift} \citep{landeiro2018conf}. Formally, confounding shift exists when two conditions are met: (i) a confounding variable $Y_c$ exists that impacts both $X$ and $Y_p$ through distributions $P(X|Y_c)$ and $P(Y_p|Y_c)$ through the backdoor path in a causal graph \citep{pearl2009causality}; (ii) a subpopulation distribution $P_{train}(Y_p|Y_c)$ is different from $P_{test}(Y_p|Y_c)$ \citep{landeiro2018conf}.

To quantitatively assess the degree of confounding shift, we use a framework proposed by \citet{ding2024backdoor} in our experiments. This allows us to perturb the target variable and confounding variable distributions in both training and test splits to different degrees through sampling from the original dataset. Under this framework, we consider a dataset with a binary target and binary confounder, the joint distribution $P(Y_p, Y_c)$  governed by the following quantity: $P(Y_c=1), P(Y_p=1), P(Y_p = 1 | Y_c = 1), P(Y_p = 1 | Y_c = 0)$. Next \citet{ding2024backdoor} introduced a positive auxiliary variable $\alpha = \frac{P(Y_p = 1 | Y_c = 1)}{P(Y_p = 1 | Y_c = 0)}$, which serves as a knob for controlling the degree of subpopulation shift. By setting different $\alpha$ values, we control the source of the positive examples. If we hold $P(Y_c=1)$ and $P(Y_p=1)$ constant, we can vary $\alpha_{train}$ and $\alpha_{test}$ to create a mixture of datasets with various degrees of shift for model evaluation. Details are described in Section~\ref{case:exp}.

\paragraph{Fairness} Fairness in machine learning seeks to ensure that models make unbiased decisions and perform equally well across different demographic groups. One widely used notion of group fairness is statistical parity, which emphasizes equal outcomes at the population level \citep{dwork2011fairness}. In binary classification tasks with a binary group attribute $G$ and binary outcome $Y$, statistical parity is measured by the absolute difference or ratio between $P(\hat{Y}=1 \mid G=1)$ and $P(\hat{Y}=1 \mid G=0)$. A smaller disparity indicates more equal treatment across groups in the model’s predictions.

Beyond statistical parity, other fairness metrics incorporate ground-truth labels to assess prediction accuracy across groups. For example, Equal Opportunity compares true positive rates between subgroups to evaluate whether the model performs equally well for individuals who belong to the positive class \citep{hardt2016equality, romano2020achieving}.

In our study setup, test set distributions can vary with the parameter $\alpha$, which governs the prevalence of dementia. This variation affects the base rate of dementia. To address this, we focus on the false positive rate (FPR), defined as $P(\hat{\text{dementia}} = 1 \mid \text{gender}, \text{dementia}=0)$—the model’s predicted dementia probability among healthy individuals. We measure the absolute difference in FPR between gender groups, denoted as $\Delta$\textit{FPR}, to assess fairness in terms of error rates.

Additionally, to evaluate whether our method reduces statistical parity, we examine the absolute output probability difference $| P(\hat{\text{dementia}} = 1 \mid \text{gender=F}) - P(\hat{\text{dementia}} = 1 \mid \text{gender=M})|$, denoted as $\Delta$\textit{SP}, on a balanced test set. This allows us to determine whether the model produces equitable predictions for male and female participants when the underlying training distribution is controlled.

\section{Experimental Setup}
\subsection{Dataset}\label{case:data}
\paragraph{DementiaBank (DB)}
The benchmark dataset used for our experiments is the Pittsburgh Corpus from DementiaBank \citep{becker1994natural, macwhinney2007talkbank} 
This corpus is a widely used resource in the fields of computational linguistics and dementia studies. It provides detailed speech and language data from elderly participants with dementia as well as healthy controls. Notably, the Pittsburgh Corpus includes responses to the \textit{Cookie Theft} picture description task from the Boston Diagnostic Aphasia Examination \citep{goodglass1983boston}. The dataset comprises 548 examples collected from longitudinal records of 290 participants. To ensure the transcripts accurately reflect the diagnosis label, we selected the last transcript for each patient as input for our model (183 female vs 107 male).

\paragraph{Carolinas Conversation Collection (CCC)} The Carolina Conversations Collection (CCC) \citep{pope2011finding} differs from DB by sourcing English conversational interviews rather than neuropsychological tasks. The corpus contains 646 interviews from 48 cognitively normal elderly individuals and 284 with dementia, with participants potentially having multiple interviews. These conversations focus on health-related autobiographical narratives and have been widely used in psycholinguistic NLP research \citep{nasreen-etal-2021-rare, li-etal-2022-gpt, farzana-parde-2023-towards}. Our study uses 394 transcripts from 70 interviewees with available gender information (323 female vs 71 male).

\subsection{Experiments} \label{case:exp}
We start by examining whether a text classification model will recognize gender confounding bias from speech data. We fine-tuned a BERT-base model \citep{devlin-etal-2019-bert} on the complete dataset and assessed its performance across gender-specific subgroups. We ran the experiments using 5-fold cross validation with 3 repeats on both the original dataset, and a perfectly balanced dataset created by down-sampling the more prevalent category. Mann-Whitney-Wilcoxon test were performed between male and female and the result is shown in Table~\ref{table:comparison} - performance discrepancies were observed among male and female examples across multiple runs. This result shows that there exists confounding by gender in the dementia detection task which is independent of the gender distribution in the dataset. This suggests that the gender of the speaker influences the language they use to complete the Cookie Theft picture description task or the phone interviews, and confound the dementia signals during model fine-tuning. Hereby, we further investigate this confounding by gender effects in dementia detection and evaluate our proposed deconfounding methods.

\begin{table}[ht!]
    \centering
    \resizebox{0.9\linewidth}{!}{ 
    \begin{tabular}{cccc}
    \toprule
    \textbf{dataset} & \textbf{setup} & \textbf{abs mean diff} & \textbf{p-value} \\
    \hline
    \multirow{2}{*}{DB} & Original & 0.055 &$<.001$ \\
    & Balanced & 0.068 & $<.001$ \\
    \hline
    \multirow{2}{*}{CCC} & Original & 0.152 &$0.002$ \\
    & Balanced & 0.102 &$0.007$\\
    \bottomrule
    \end{tabular}
    }
    \caption{Two sided Mann-Whitney-Wilcoxon test results of male and female dementia prediction performance (AUPRC) across different setups.}
    \label{table:comparison}
\end{table}

\begin{figure*}[htbp]
    \centering
    \includegraphics[width=\linewidth]{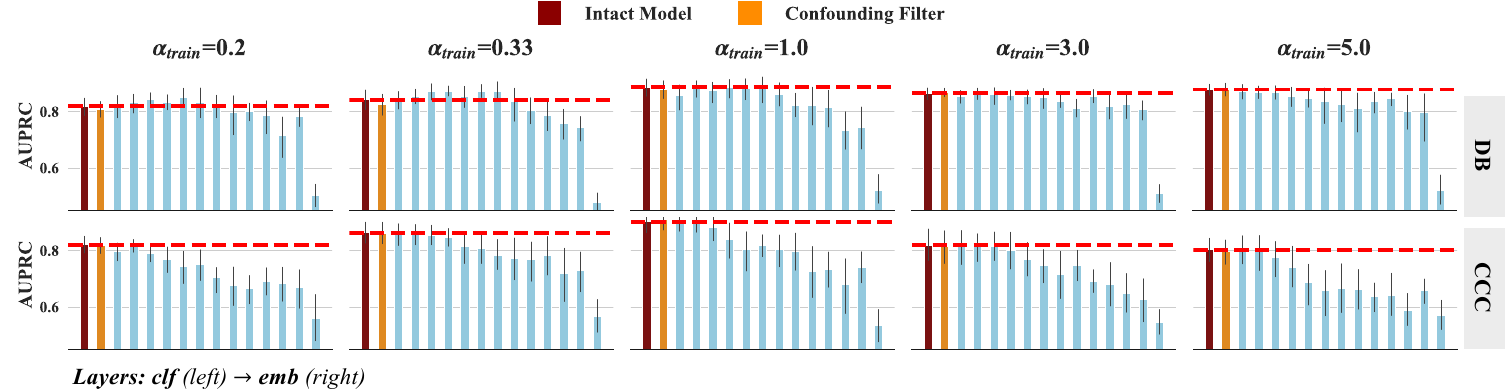}
    \caption{ECF filtering with 15\% masking rate across different confounding shifts from two dataset.}
    \label{fig:aps_ecf}
\end{figure*}

\paragraph{Dataset Perturbation}
As described in Section \ref{sec:evalmethod}
, we manipulated the conditional distribution of dementia by gender in our dataset through random sampling, creating a series of datasets with varying levels of confounding shift. In our experiments, dementia cases and female participants are coded as $1$, respectively. We fixed $P(\text{gender} = 1) = 0.5$ and $P(\text{dementia} = 1) = 0.5$ in both the training and test sets to ensure fair comparisons across different configurations. This way, the dataset is balanced with respect to both dementia and gender. Then we adjusted the value of $\alpha = \frac{P(\text{dementia}|\text{female})}{P(\text{dementia}|\text{male})}$ to create an imbalance in the source of dementia cases (subpopulation shift). If $\alpha > 1$, more dementia cases are drawn from females, while $\alpha < 1$ indicates the opposite. The further $\alpha$ is from $1$, the more severe the imbalance. To evaluate the model's robustness to confounding shifts, the model is trained on one $\alpha_{\text{train}}$ value and tested on its reciprocal value $\alpha_{\text{test}} = \frac{1}{\alpha_{\text{train}}}$, simulating an extreme shift in the test set compared to the distribution the model was exposed to during training. The selection of $\frac{1}{\alpha_{\text{train}}}$ is entirely arbitrary and is intended solely to illustrate the shift magnitude and establish the testbed for evaluation. Models are trained for 20 epochs on 480 training examples, validated on 120 examples and evaluated on 150 examples for each configuration. Among them, the training set and validation set are sampled from $\alpha$, while the test set is sourced from $\frac{1}{\alpha}$. The best checkpoint is selected based on the AUPRC on the validation set, using early stopping to prevent overfitting.

\paragraph{Extended Confounding Filter}
The encoder model we used for dementia detection is BERT-base, with 12 encoder layers and 12 attention heads in each layer. Once we obtain the dementia fine-tuned model $f(x)$ after the first Phase, we take a snapshot of the parameters and only make some parts of it trainable towards the gender label in the second Phase. The trainable layers start with {cls}, and one layer is sequentially added to the trainable set. Eventually, the trainable set becomes $\{cls, layer12, layer11, ..., layer1, emb \}$ and spans the whole network. Then for each trainable set, $f_d$ is trained toward gender prediction. We ranked the weights that changed in each layer and selected the top 15\% of the weights with the most significant changes in each layer to mask (Figure~\ref{fig:exconf}), following the approach in \citet{wang2019removing}. Then we evaluated the masked models. We include results of choosing different masking ratios for ECF in Appendix \ref{appendix:ecf_ratios}.
\vspace{-1mm}

\paragraph{Dual Filter}
In the Dual Filter approach, we track the global weight change throughout the model's architecture. The classification head is exempt from tracking as it is training toward two different tasks and the weights in the classification head are assumed to have the most significant change compared to the rest of network. We first obtain two lists of weights change matrices from $f(x)$ and $g(x)$, using the same approach as Extended CF. Then we rank and select the top \textit{k}\% weights by their locations in the network. A sequence of \textit{k} values are tested, ranging from $0$ to $60$ and step size of 1. Then three sets ($M_I, M_D, M_I\cap M_D$) are calculated and applied to $f(x)$ to create the masked model. Note when training toward gender in both Extended CF and Dual Filter, we select only \textit{non-dementia} cases to let the model learn from texts that are representative of the gender differences. Consequently, only healthy cases are used in the evaluation.

\section{Results}
\vspace{-1mm}

\subsection{Extended Confounding Filter}

Figure~\ref{fig:aps_ecf} presents the results of the Extended Confounding Filter (ECF). The red dotted line represents the performance of the intact model, while the bars illustrate models where weights are progressively eliminated, layer by layer, from left to right until the embedding layer is reached (the rightmost bar). The orange bar represents the original Confounding Filter approach, where only the classification head is trained in the second phase and then masked.

We observe that simply applying the Confounding Filter to the classification layer is insufficient to mitigate confounding bias. Propagating masks layer by layer helps maintain or even improve dementia classification performance. Specifically, the model remains resilient in dementia detection when gender-associated weights are removed from the upper layers, with no significant performance drops occurring until weights are ablated from lower layers. This aligns with prior work \citep{li-etal-2024-big}, which reports similar robustness in linguistic feature encoding. Interestingly, in some cases, removing gender-associated weights from certain layers improves AUPRC compared to the intact model, suggesting potential patterns that warrant further investigation.
Across both datasets, a “ladder” effect emerges due to confounding shift: models trained and tested on the same distribution ($\alpha = 1$) achieve the highest performance, while the performance declines as $\alpha_{train}$ deviates from 1.

In particular, the token embedding layer plays a critical role in dementia detection—our experiments show that removing even a small fraction of its weights drastically alters the model’s performance. Further research is needed to better understand its contribution when using transformer models for dementia detection.
\vspace{-1mm}

\subsection{Dual Filter}

\vspace{-2mm}

\begin{figure}[ht!]
    \centering
    \includegraphics[clip=true, trim={8 0 0 0}, width=1.\linewidth]{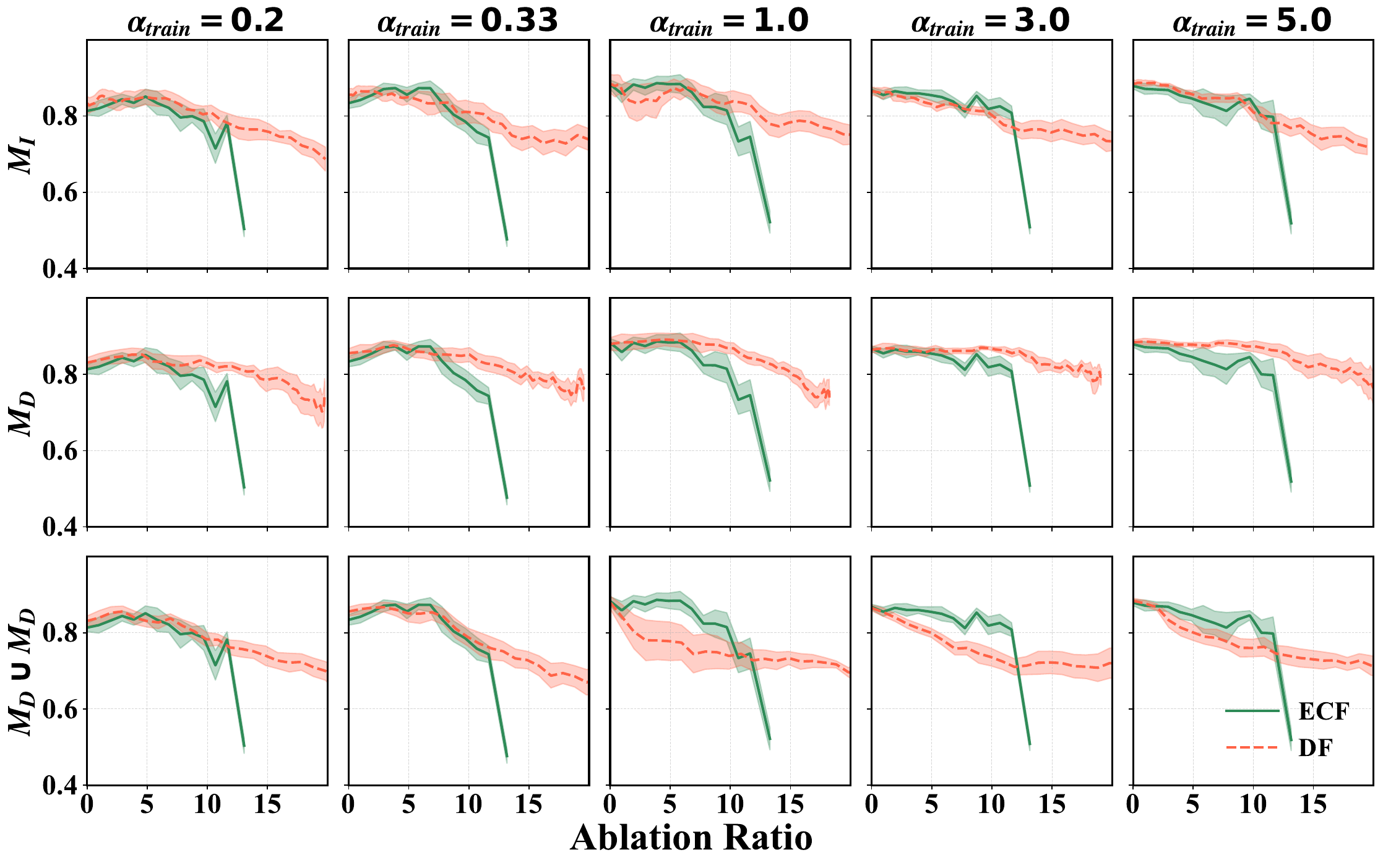}
    \caption{AUPRC (y-axis) on ECF and DF for different $\alpha_{train}$ configurations. The x-axis represents the \% of ablated weights through the whole network.}
    \vspace{-3mm}
    \label{fig:dualfilter}
\end{figure}
In Figure~\ref{fig:dualfilter}, we visualize the dementia prediction performance change on the DB dataset as we apply three different types of mask to the original model and gradually increase the masking ratio. The results from ECF with 15\% layer-specific masking ratio are added for comparison. The plot shows the relation between how many weight entries are ablated within the whole network against model AUPRC. The rows indicate three types of masks that are generated by Dual Filter and the columns indicate the specific $\alpha_{train}$ configurations that control the distribution shift.

\begin{figure}[ht!]
    \centering
    \includegraphics[clip=true, trim={8 0 0 0}, width=1.\linewidth]{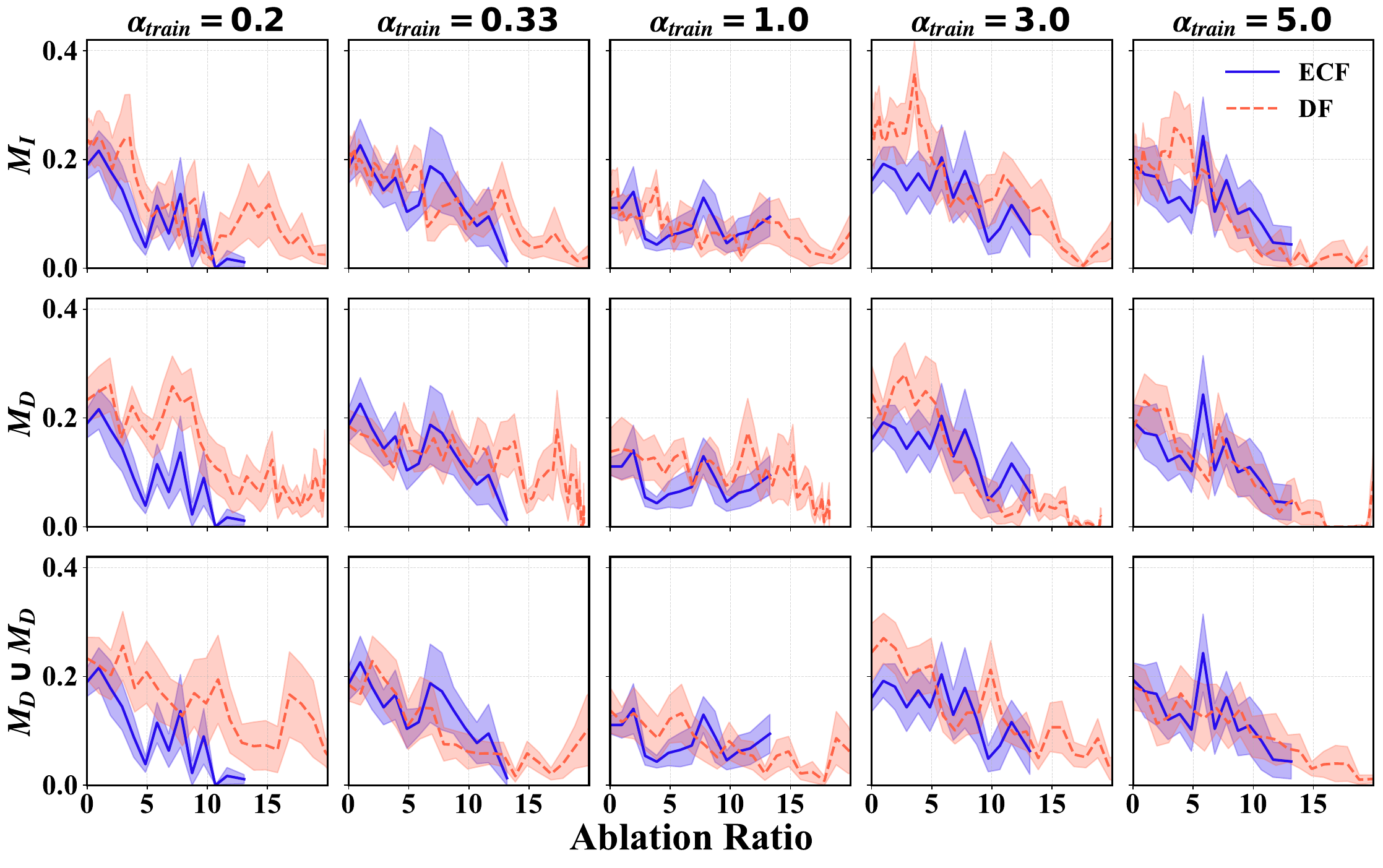}
    \caption{$\Delta$\textit{FPR} (y-axis) on ECF and DF for different $\alpha_{train}$ configurations.}
    \vspace{-3mm}
    \label{fig:dualfilterfpr}
\end{figure}

Next, we show the absolute False Positive Rate difference ($\Delta$\textit{FPR}) between females and males calculated under both ECF and DF methods. Figure~\ref{fig:dualfilterfpr} shows the change in FPR gaps as the ablation ratios increase for all three types of masks. The mask type is indicated in rows while the columns represent different $\alpha_{train}$.
Similar trends are also observed in the CCC dataset, shown in Appendix~\ref{sec:ccc_df}.

\begin{figure}[ht!]
    \centering
    \includegraphics[clip=true, trim={8 0 0 0}, width=1.\linewidth]{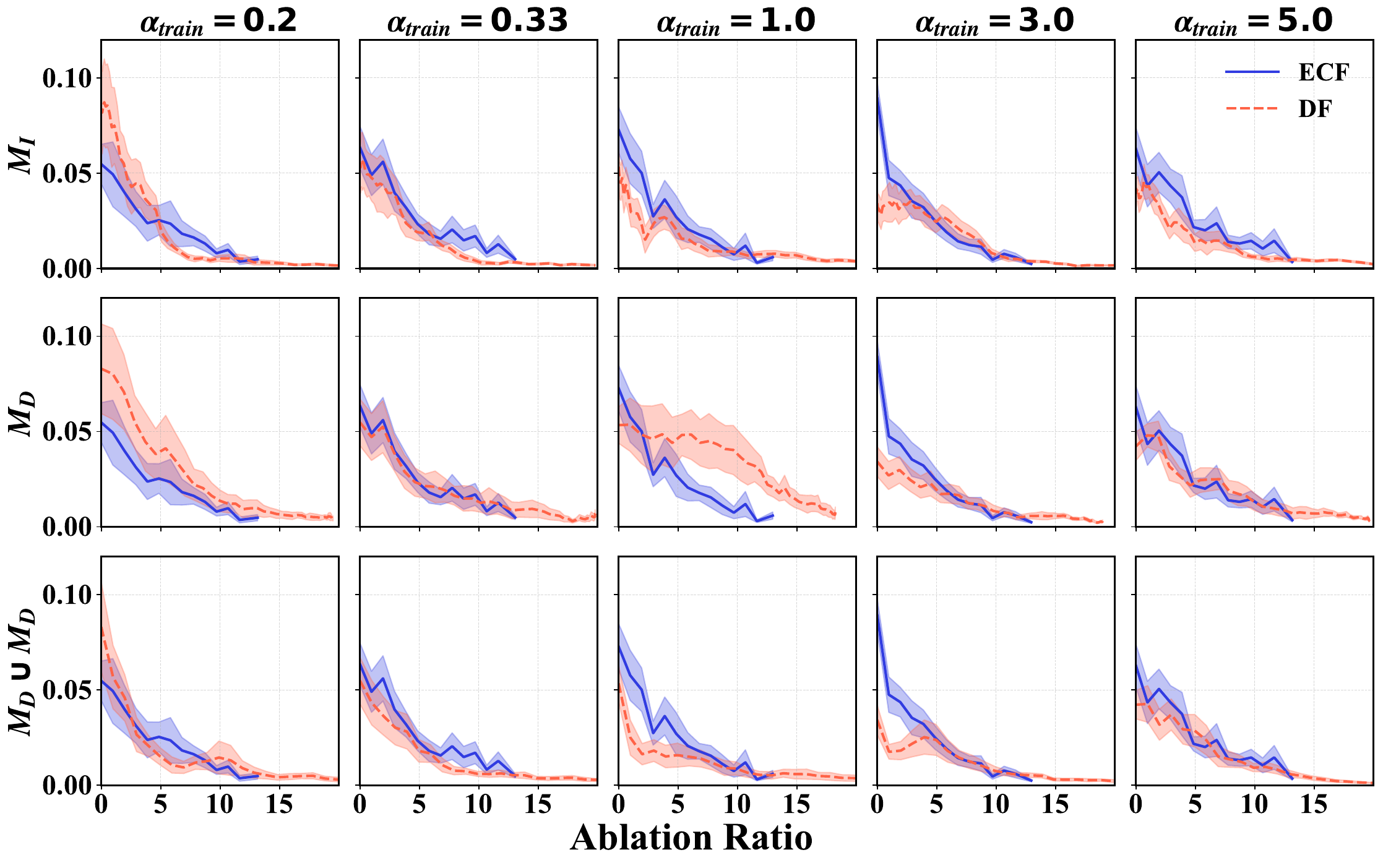}
    \caption{$\Delta$\textit{SP} (y-axis) on ECF and DF for different $\alpha_{train}$ configurations.}
    \label{fig:dualfiltersp}
\end{figure}

\begin{figure*}[htbp!]
    \centering
    \includegraphics[ width=\linewidth]{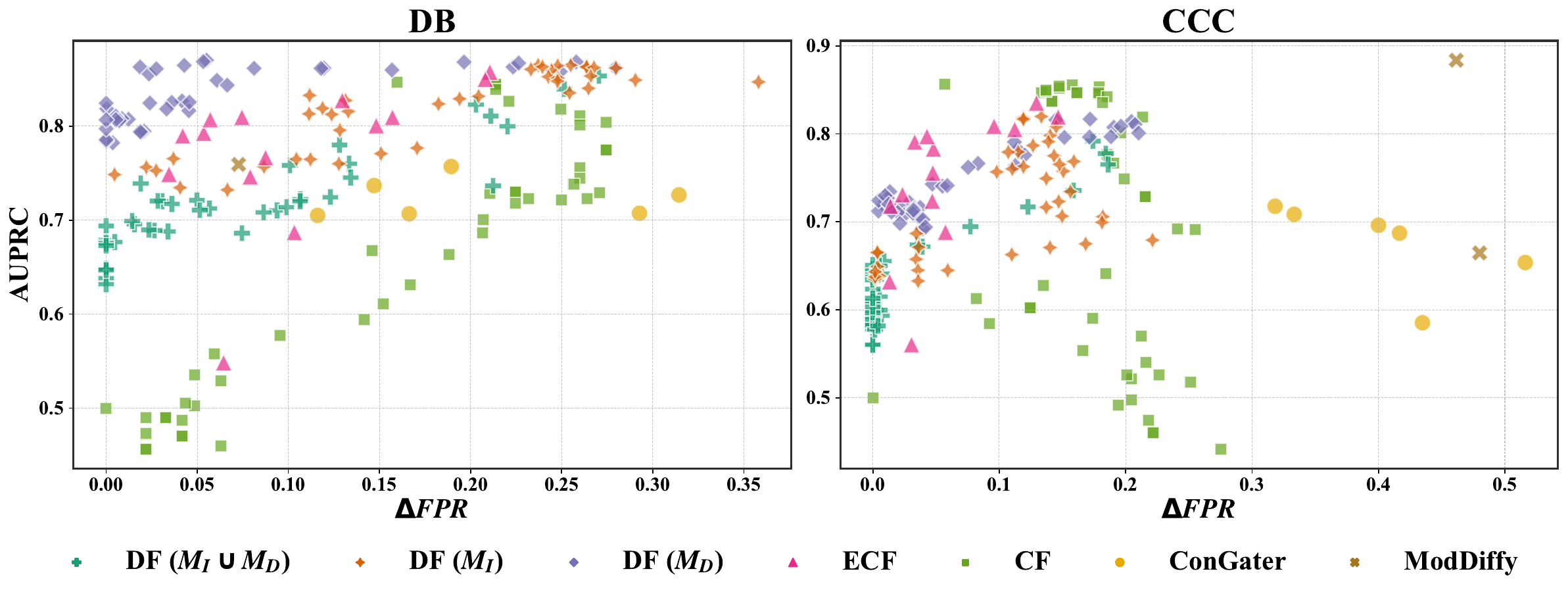}
        \vspace{-7mm}
    \caption{AUPRC vs. $\Delta$\textit{FPR} among different methods for two datasets.}
    \label{fig:baseline}
    \vspace{-5mm}
\end{figure*}
The results in Figure~\ref{fig:dualfilter} show that all three masking strategies display similar trends under varying confounding shift configurations. Both the $M_I$ and $M_D$ masks exhibit stronger resilience across different ablation ratios compared to the ECF baseline. In contrast, the union mask ($M_I \cup M_D$), which removes a larger set of weights, exhibits comparable or slightly reduced resilience compared to ECF across different $\alpha$ settings. This suggests that removing all gender-related weights without considering their association with dementia labels impair model performance under distributional shifts.

Paired with Figure~\ref{fig:dualfilterfpr}, we observe that in certain intervals of the ablation ratios, the performance remains stable while the fairness metric improves. For example, at $\alpha = 0.2$ in the DB dataset, removing 10\% of the weights from $M_I$ (the intersection mask) achieves model AUPRC at 0.80, which drops only slightly from the original model (0.83), while the FPR difference between male and female drops from 0.23 to 0.03. The trade-offs between AUPRC and $\Delta$\textit{FPR} suggest an entanglement between the weights responsible for dementia detection and those associated with gender, particularly in weight entries that undergo the most change across different layers of the network. We further analyze this entanglement across layers under different confounding shift settings in Appendix~\ref{entangle}. Figure~\ref{fig:dualfilterfpr} also suggests that FPR gaps are more severe in the original model under extreme confounding shifts.

In addition, we assess model fairness on the DB dataset using statistical parity, with the evaluation conducted on a balanced test set ($\alpha = 1$). As shown in Figure~\ref{fig:dualfiltersp}, both proposed weight masking methods substantially reduce the prediction disparity between male and female participants across different training data shifts. This improvement also suggests that the methods successfully identify and suppress gender-related weights in the model, leading to more equitable predictions.

\subsection{Comparisons with other methods}
We train different models on data with an arbitrarily selected setting with $\alpha_{train} = 3$ to represent a confounding shift case. That means the dementia cases for training are sourced three times more from the female cohort than from the male cohort. We then test on samples from the configuration of $\alpha_{test} = \frac{1}{3}$. We evaluate the results on ECF, DF and other baselines using the AUPRC-$\Delta$\textit{FPR} curve (Figure~\ref{fig:baseline}), in which the upper-left points represent an ideal classifier, with both high accuracy and fairness. 

Our experiments show that ECF achieves the best trade-off on the CCC dataset, while DF ($M_D$) outperforms the other methods on the DB dataset. Across both datasets, our methods consistently outperform adapter-based baselines and naive Confounding Filters, demonstrating their effectiveness in mitigating confounding shifts. Notably, compared to naive Confounding Filters, our methods achieve a more favorable trade-off, maintaining higher AUPRC at the several FPR disparity levels. 

Additionally, we note that weight masking approaches offer a more fine-grained trade-off trajectory than loss-optimization methods, providing greater flexibility across various use cases.

\section{Discussion}
\vspace{-1mm}
 From the experiments, we conclude that both ECF and DF effectively mask gender-related weights within a BERT-base model, improving gender parity in outcomes while maintaining comparable performance in dementia detection under various degrees of confounding shifts. Comparisons with existing baselines demonstrate that our proposed methods achieve a more favorable trade-off between model performance and gender parity. We further discuss the broader impact of the work. 
\paragraph{Clinical Implication} While this work exclusively focuses on the gender-confounding issue in dementia, its broader implications extend to other medical AI applications. Unbalanced data is common in many medical settings, and inductive bias in deep learning models can lead to  misdiagnoses or uneven treatment recommendations during model inference. Our findings emphasize the need for proactive bias mitigation strategies in low-resource domains like dementia research, and future work should explore extending these deconfounding methods to additional clinical variables.
\paragraph{Generalizability} Both the ECF and DF methods are model-agnostic and can be applied to any transformer-based architecture. Additionally, both methods can be adapted for non-binary confounders by formulating them as a multi-class classification task during model fine-tuning. Although this paper focuses on a single clinical task with a binary target and a binary confounder,the proposed framework can be extended to more complex confounding scenarios.
\paragraph{Scalability} ECF suffers from scalability issues due to its retraining process at each layer to exploit the trade-offs, while DF is more computationally efficient, requiring only two fine-tuning steps. Therefore, its computational complexity scales linearly with the size of the dataset, making it more practical for larger datasets and models.


\section{Conclusion}
\vspace{-1mm}
In this paper, we address gender confounding bias learned during model fine-tuning and propose two model-agnostic methods for filtering confounder-associated weights in transformer neural networks. We apply these methods to dementia detection tasks, demonstrating their potential utility in clinical practice. Our findings indicate that unaddressed confounding shifts can degrade model performance even when the overall label and group distributions are balanced. Experimental results compare the identification of gender-associated weights both layer-wise and across the entire model. Both methods show minimally degraded performance on the dementia detection task while reducing gender bias. We observed non-monotonic responses across layers, suggesting further investigation is needed to understand the inner workings of even small transformer models. Lastly, we note that ensuring fairness and maintaining model performance often involve trade-offs, and real-world decisions should consider multiple factors, including bias tolerance and use-case specifics.

\section*{Limitations}
\paragraph{Dataset} The experiments of our proposed methods are only conducted on two relatively small datasets; therefore, generalizability to other datasets will need to be further investigated. In addition, given the small data size, manifesting different levels of confounding shift requires repetitive sampling to meet the desired subgroup distribution. Thus the resultant dataset contains a significant amount of duplicates that may reduce the strength of the findings and would need to be replicated in larger datasets.
\paragraph{Methods} In ECF methods, even though the approach we take is the most straightforward and allows the model to absorb unidirectional effects, we ignore the possibility of other combinations of layer freezing inside the network and leave it to be explored in future work.
\paragraph{Experiments} While we acknowledge BERT-base as a good starting point for investigation, we did not include other encoder models in this work. Also, while we briefly discussed some other weight importance measurements to isolate weights that impact certain outputs, we did not include and compare them with our current approach for de-confounding mainly due to the publicly available code for other prior work not being model-agnostic.
\section*{Ethical Statement}

\paragraph{Dataset Privacy} The two datasets utilized in this study are publicly available (upon request) and have been fully de-identified.
1) DementiaBank transcripts: These are collected from picture description tasks in cognitive tests designed to assess structural language skills.
2) Carolina Conversations Collection transcripts: These are derived from interviews about life stories or personal well-being.
While these datasets include demographic information about the subjects, they are considered a health-related speech dataset instead of medical dataset in general sense, and the privacy concerns associated with their use are minimal. 

\paragraph{Potential Downstream Risk} While the deployment of model-based dementia screening tools has the potential to support timely interventions and improve patient outcomes through early detection of cognitive decline, these benefits must be carefully weighed against ethical and practical concerns. For example, false positives of model predictions may cause distress and anxiety to patients or lead to premature clinical decisions; Such models can also be applied for non-medical purposes such as monitoring cognitive status or unregulated cognitive assessments of vulnerable populations, which may aggravate discrimination against those groups.

\paragraph{Gender Clarification} The datasets used in this study include only participants identified as male or female. In the DB dataset, gender labels are inferred from participants’ reported biological sex, whereas in the CCC dataset, gender is explicitly provided in the metadata. We acknowledge that equating gender with biological sex in the DB dataset is a strong assumption, and this limitation warrants further investigation.

\section*{Acknowledgment}
This work was supported by U.S. National Library of Medicine Grant (R01LM014056).
\bibliography{custom}

\newpage

\appendix
\setcounter{section}{0}
\section{Evidence of Confounding}
\renewcommand{\thefigure}{S\arabic{figure}}
\setcounter{figure}{0}
\setcounter{table}{0}
\renewcommand{\thetable}{S\arabic{table}}
\subsection{Gender bias in dementia detection}
\begin{figure}[ht!]
    \centering
    \begin{subfigure}[b]{0.48\linewidth} 
        \centering
        \includegraphics[width=\linewidth]{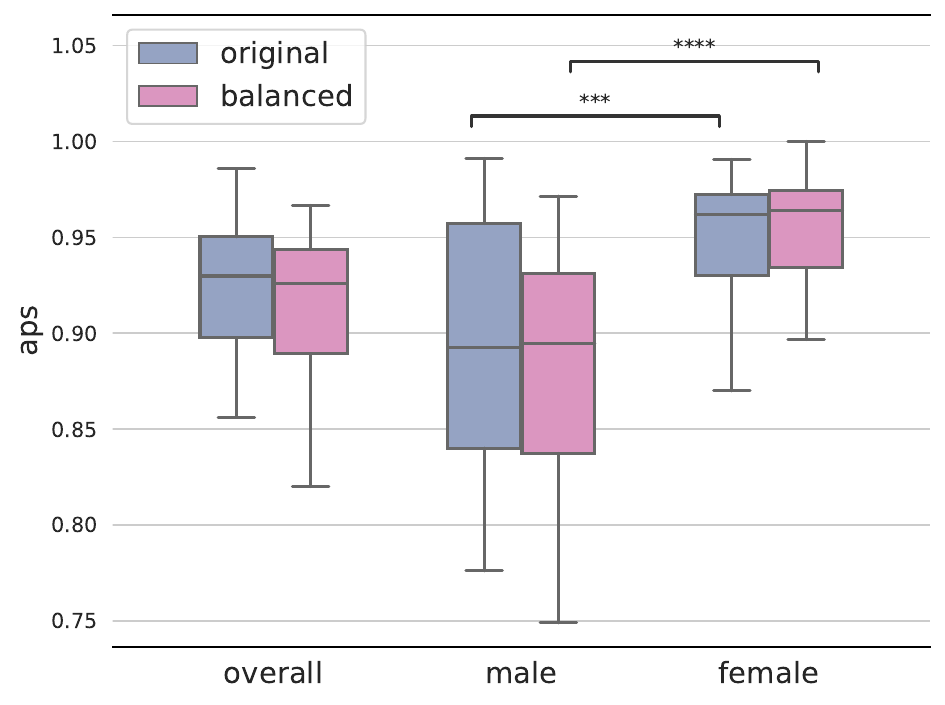}
        \caption{AUPRC results from DB}
        \label{pitts_raw}
    \end{subfigure}
    \hfill
    \begin{subfigure}[b]{0.48\linewidth} 
        \centering
        \includegraphics[width=\linewidth]{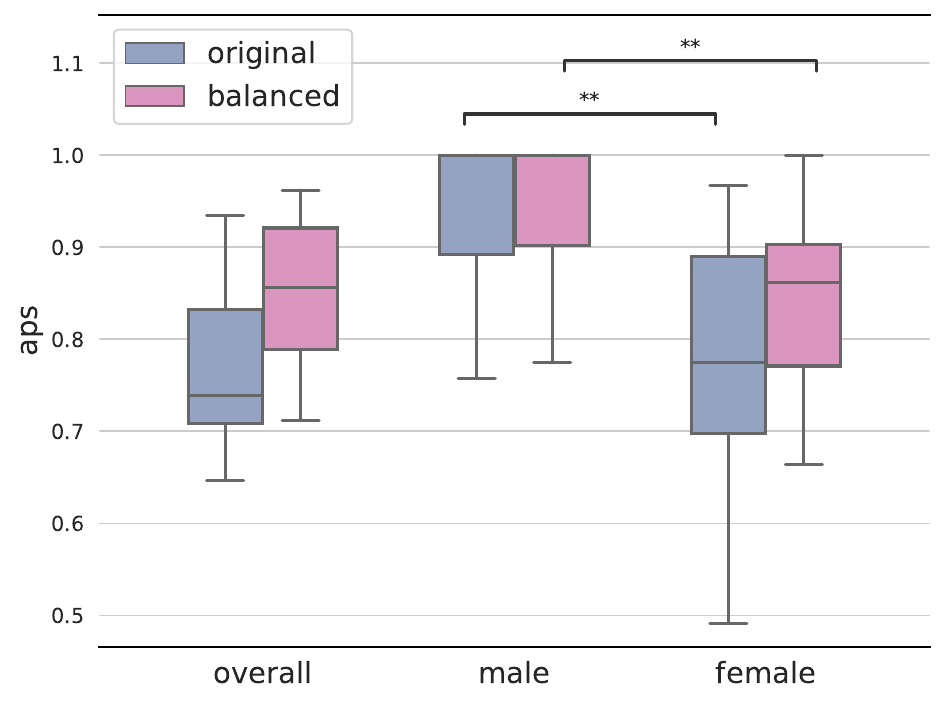}
        \caption{AUPRC results from CCC}
        \label{ccc_raw}
    \end{subfigure}
    \caption{Performance discrepancy between male and female in two datasets using the BERT-base model.}
    \label{fig:rawdisparity}
\end{figure}

\vspace{-5mm}
\subsection{Dataset Statistics}\label{data_dist}
\begin{table}[ht!]
\centering
\begin{minipage}{0.45\linewidth}
\centering
\resizebox{\linewidth}{!}{ 
\begin{tabular}{ccc}
\toprule
\multicolumn{3}{c}{DB Dataset} \\
\midrule
Gender & Label & Count \\
\midrule
Female & 0 & 57  \\
\rowcolor{gray!10}Female & 1 & 126 \\
Male   & 0 & 41  \\
\rowcolor{gray!10}Male   & 1 & 66  \\
\bottomrule
\end{tabular}
}
\end{minipage}
\hspace{0.05\linewidth} 
\begin{minipage}{0.45\linewidth}
\centering
\resizebox{\linewidth}{!}{ 
\begin{tabular}{ccc}
\toprule
\multicolumn{3}{c}{CCC Dataset} \\
\midrule
Gender & Label & Count \\
\midrule
Female & 0 & 220 \\
\rowcolor{gray!10}Female & 1 & 103 \\
Male & 0 & 42  \\
\rowcolor{gray!10}Male & 1 & 29  \\
\bottomrule
\end{tabular}
}
\end{minipage}
\caption{DB and CCC Datasets gender-label counts}
\end{table}
\vspace{-5mm}
\section{Experiments Setup}
\subsection{Finetuning hyperparameters}
\begin{table}[h]
    \centering
    \resizebox{\linewidth}{!}{%
    \begin{tabular}{llr}
        \toprule
        \textbf{} & \textbf{parameter} & \textbf{value} \\
        \midrule
        \textbf{Data Size} & Train & 480 \\
                          & Validation & 120 \\
                          & Test & 150 \\
        \midrule
        \textbf{Hyperparameters} & Max sequence length & 256 \\
                                & Number of epochs & 20 \\
                                & Early stopping tolerance & 5 \\
                                & Early stopping metric & AUPRC \\
                                & Optimizer & AdamW \\
                                & Scheduler & Linear \\
                                & Warm up steps & 50 \\
                                & Learning rate & 1e-5 \\
        \bottomrule
    \end{tabular}
    }
    \caption{Model and Data Configurations for Finetuning}
    \label{tab:data_hyperparams}
\end{table}
\vspace{-5mm}
\subsection{Hyperparameter selection for baseline methods}
In Table~\ref{tab:baseline_hyp}, we present the selected hyperparameters for the two adapter-based baseline methods along with their corresponding definitions.
\begin{table*}[ht!]
    \centering
    \small
    \resizebox{0.65\linewidth}{!}{
    \begin{tabular}{cccc}
    \toprule
        Methods & Hyparam & Values & Note\\
        \midrule
         \textsc{Congater}& $\omega $ & $0.2, 0.4, 0.6, 0.8, 1.0$ & \textit{gate sensitivity} \\
         \midrule
         \textsc{Moddiffy} & \texttt{fixmask\_pct} & $0.1, 0.3, 0.5, 0.7, 0.9$ & \textit{weight mask cutoffs} \\
         \bottomrule
    \end{tabular}
    }
    \caption{Hyperparameters used in baseline experiments}
    \label{tab:baseline_hyp}
\end{table*}

\section{Additional Results}

\subsection{CF vs ECF on CCC dataset}\label{sec:ccc_df}
\begin{figure}[ht!]
    \centering
    \includegraphics[width=1\linewidth]{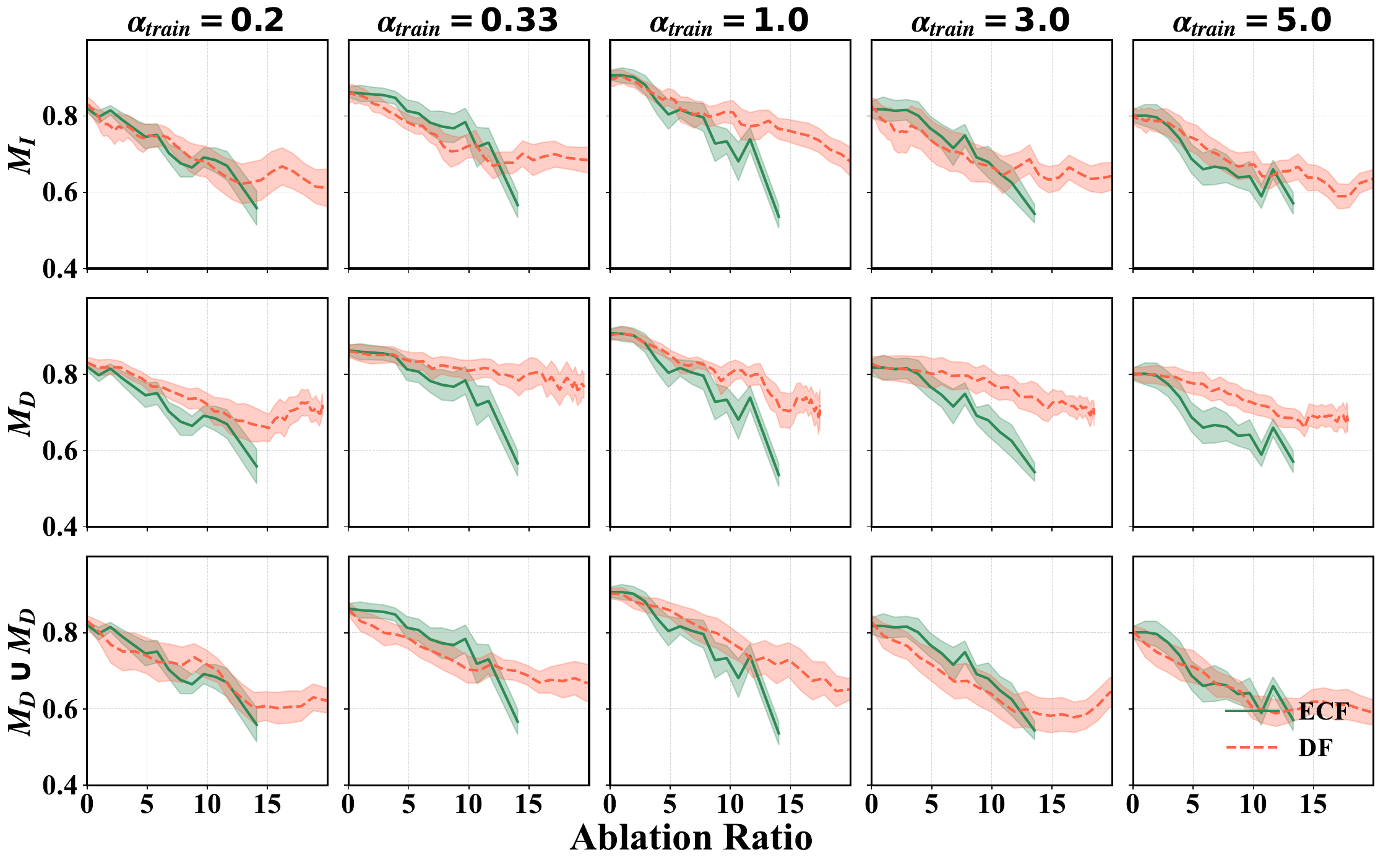}
    \caption{Relationship between AUPRC and weights ablation on CCC dataset for ECF and DF.}
    \vspace{-5mm}
    \label{fig:ccc_df}
\end{figure}

\begin{figure}[ht!]
    \centering
    \includegraphics[width=1\linewidth]{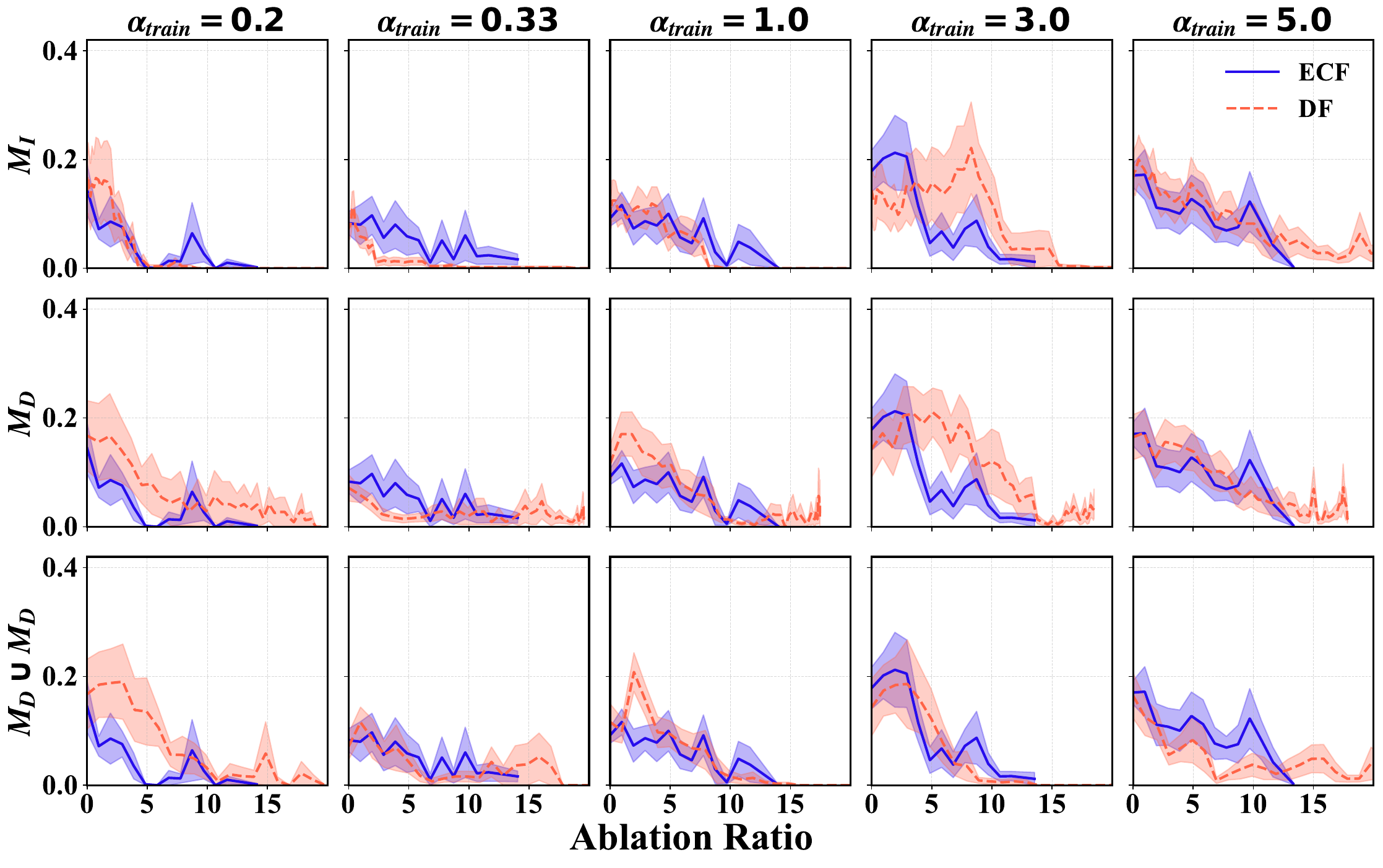}
    \caption{Relationship between $\Delta$\textit{FPR} and weights ablation on CCC dataset for ECF and DF.}
    \vspace{-5mm}
    \label{fig:ccc_fpr}
\end{figure}

In this section we show the results for EF and DF comparisons for AUPRC and $\Delta$\textit{FPR} against the ablation ratio. Figure~\ref{fig:ccc_df} demonstrates the ablation effects on the AUPRC metric and Figure~\ref{fig:ccc_fpr} shows the effect for the absolute FPR difference between female and male. The results suggest both methods work effectively on the CCC dataset.

\subsection{ECF with different masking ratios}\label{appendix:ecf_ratios}
\begin{figure*}[ht!]
    \centering
    \includegraphics[width=1\linewidth]{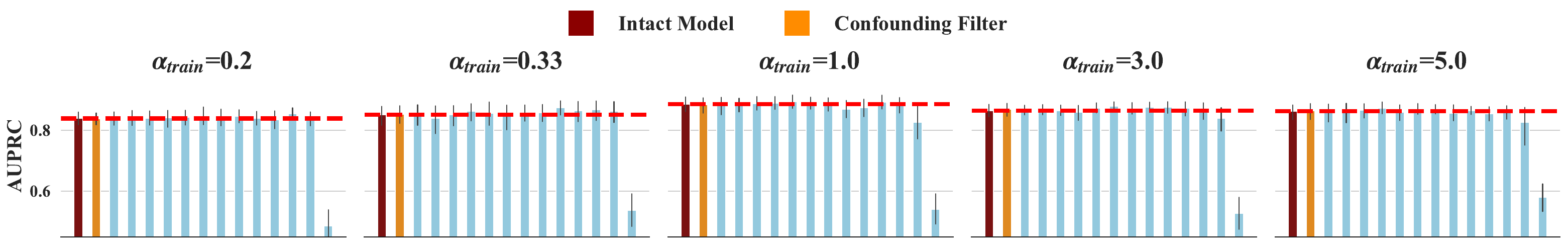}
    \caption{ECF filtering with 5\% masking rate across different confounding shifts on DB dataset.}
    \vspace{-3mm}
    \label{fig:ecf_5}
\end{figure*}

\begin{figure*}[ht!]
    \centering
    \includegraphics[width=1\linewidth]{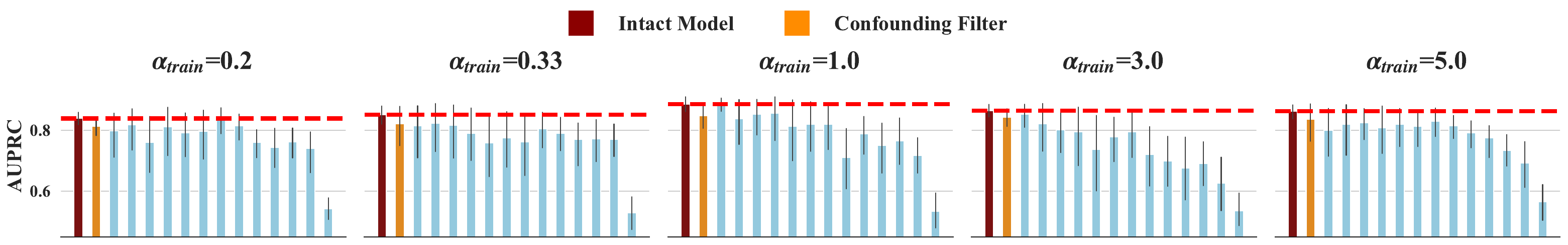}
    \caption{ECF filtering with 25\% masking rate across different confounding shifts on DB dataset.}
    \vspace{-3mm}
    \label{fig:ecf_25}
\end{figure*}

\begin{figure*}[ht!]
    \centering
    \includegraphics[width=1\linewidth]{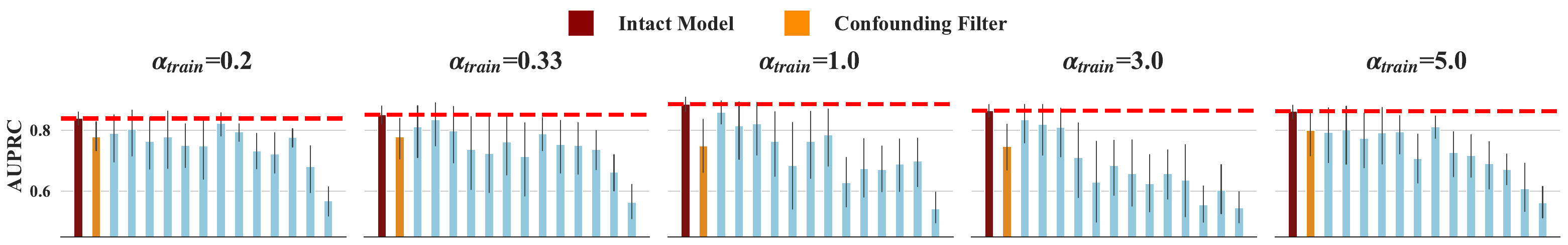}
    \caption{ECF filtering with 35\% masking rate across different confounding shifts on DB dataset.}
    \vspace{-3mm}
    \label{fig:ecf_35}
\end{figure*}

\begin{figure}[ht!]
    \centering
    \includegraphics[width=1\linewidth]{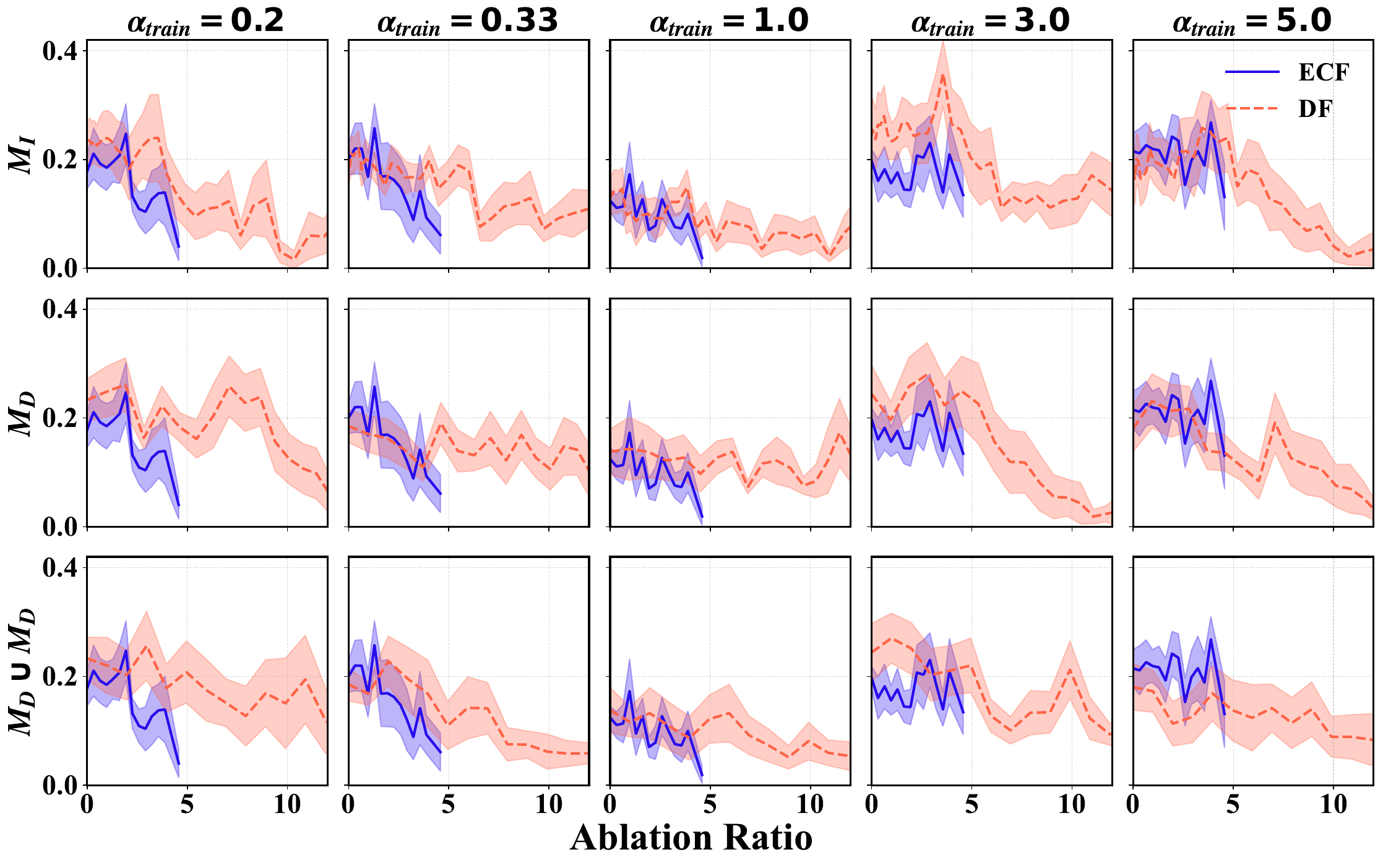}
    \caption{Comparison of ECF with 5\% masking ratio and DF on $\Delta$\textit{FPR} against ablation ratio on DB dataset.}
    \label{fig:ecf_5_fpr}
\end{figure}

In this section, we present ECF results with different masking ratios (5\%, 25\% and 35\%). Figrue~\ref{fig:ecf_5}, \ref{fig:ecf_25}, and \ref{fig:ecf_35} demonstrate the results. We can observe that as the masking ratio increases, the model performance on dementia detection regarding AUPRC drops significantly after several layers. We then assess whether masking only 5\% of the weights in ECF can effectively reduce FPR gaps. As shown in Figure~\ref{fig:ecf_5_fpr}, substantial fairness improvements can be achieved by removing a small fraction of weights from each layer of the BERT-base model under certain configurations. This suggests that the optimal masking ratio may vary depending on the dataset and experimental setup.

\section{Analysis}
\subsection{Relationship of three types of masks in Dual Filter}
\begin{figure}[ht!]
    \centering
    \includegraphics[width=0.6\linewidth]{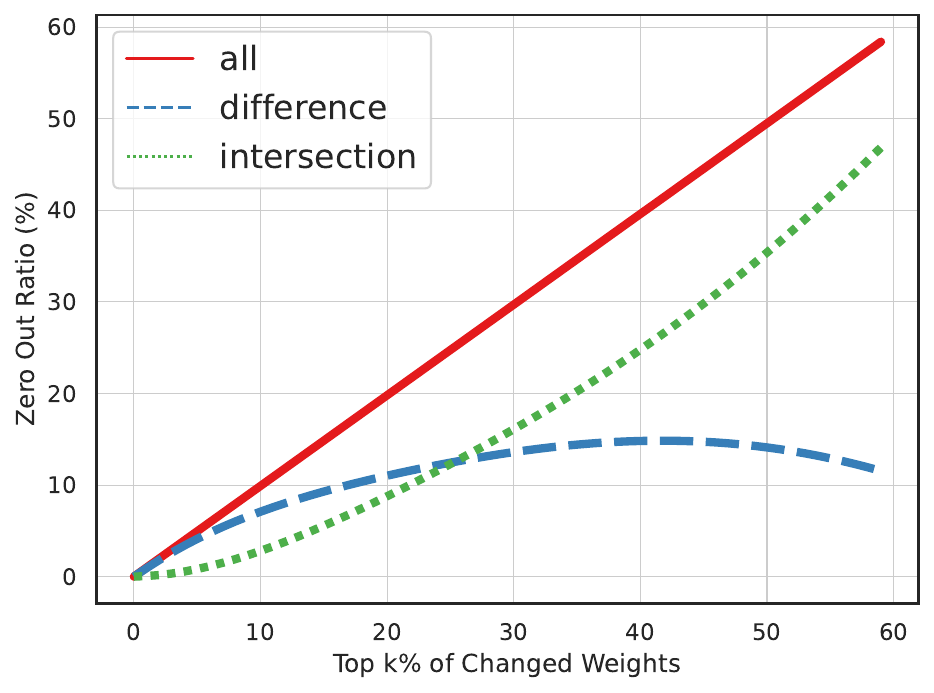}
    \caption{Ablation ratio by each mask against the total masking ratio}
    \label{fig:mask_types}
\end{figure}

The relationships between the ablation ratio of the three types of masks and the choice of $k$ are shown in Figure \ref{fig:mask_types}. As we tune $k$ to increase the coverage of active parameters in the model, the size of $M_D$ first grows then reaches its peak at around $k = 40$ and then falls back to zero, while the size of $M_I$ keeps increasing. 
\subsection{Entanglement Analysis}\label{entangle}
\begin{figure*}[ht!]
    \centering
    \includegraphics[width=0.65\linewidth]{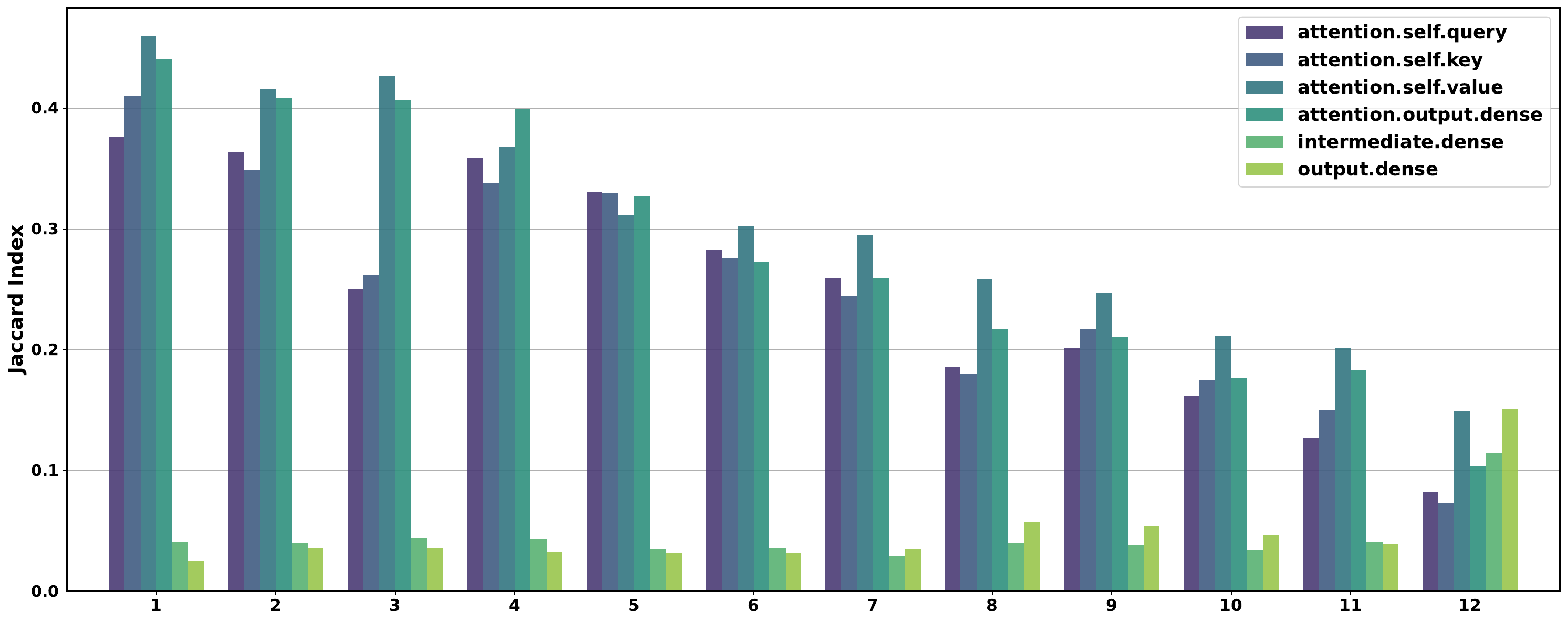}
    \caption{Jaccard Index for each of the tracked matrix in Dual Filter ($\alpha_{train} = 0.20, \alpha_{test} = 5.0$)}
    \label{fig:jaccard-0.2}
\end{figure*}

\begin{figure*}[ht!]
    \centering
    \includegraphics[width=0.65\linewidth]{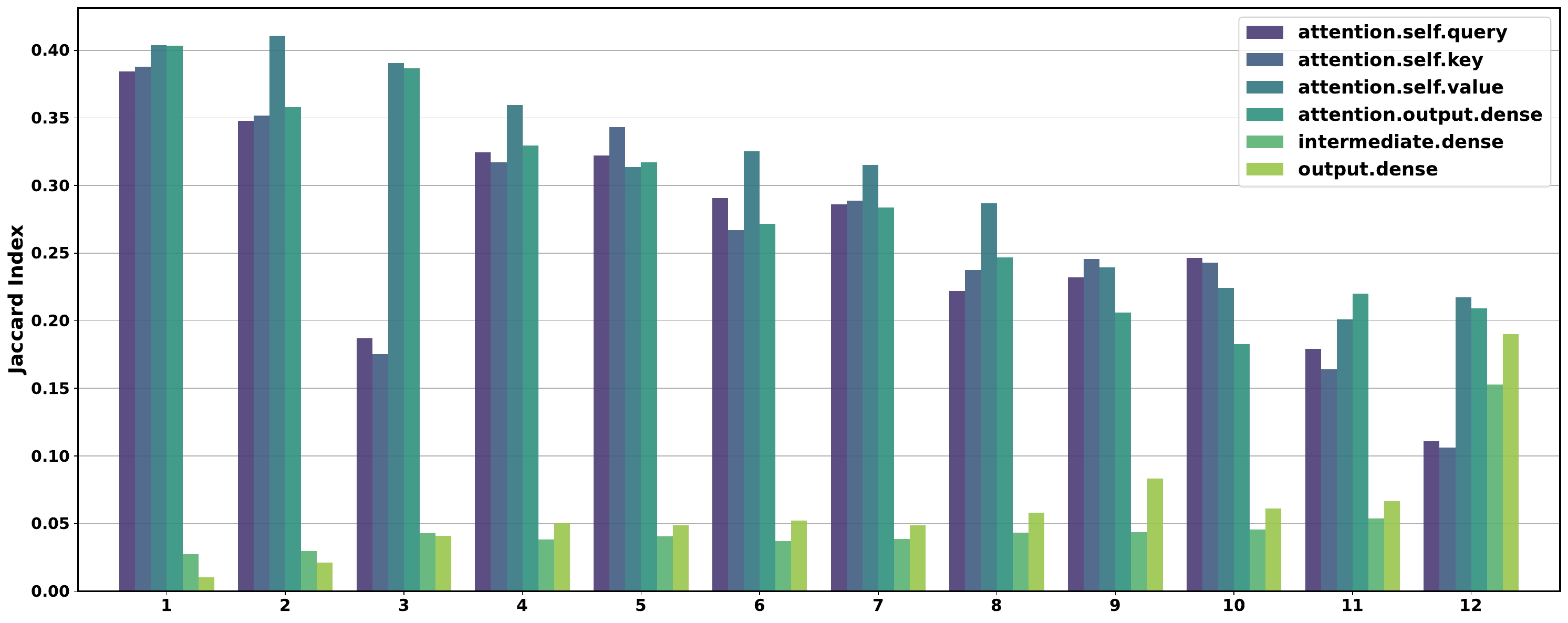}
    \caption{Jaccard Index for each of the tracked matrix in Dual Filter ($\alpha_{train} = 0.33, \alpha_{test} = 3.0$)}
    \label{fig:jaccard-0.33}
\end{figure*}

\begin{figure*}[ht!]
    \centering
    \includegraphics[width=0.65\linewidth]{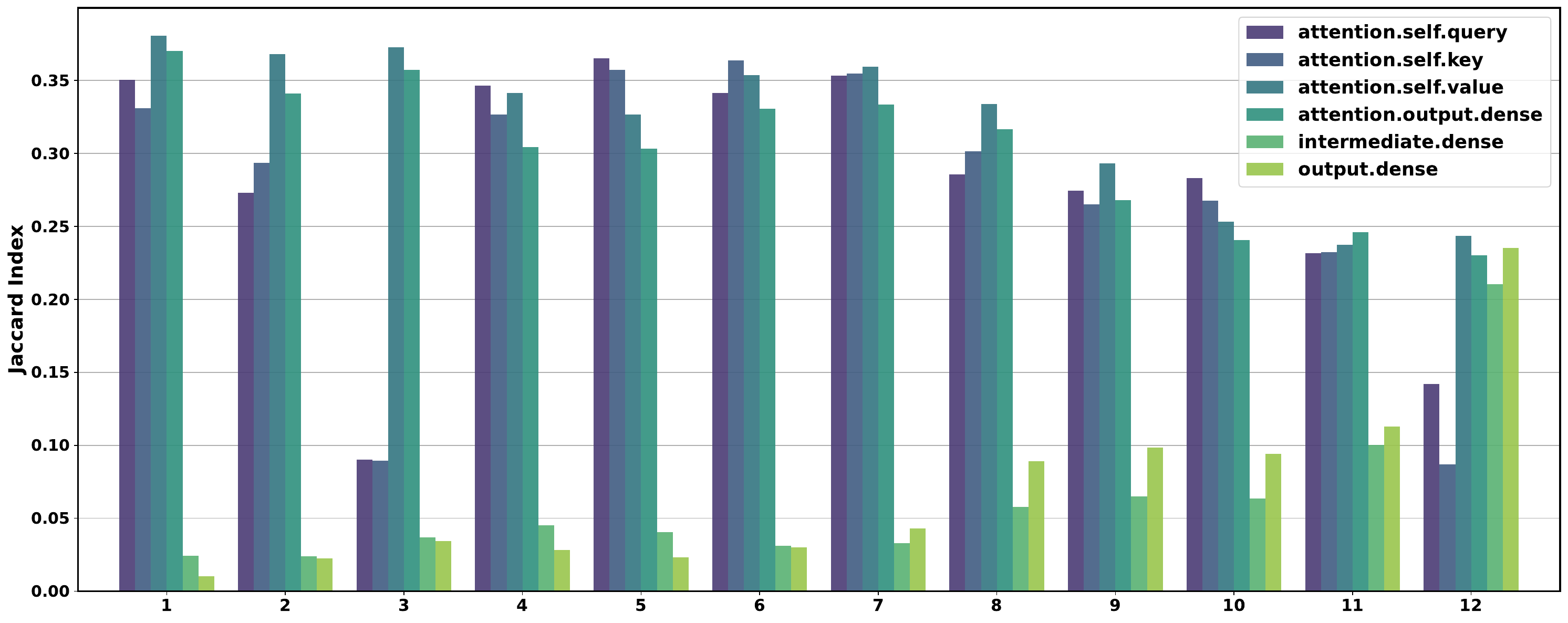}
    \caption{Jaccard Index for each of the tracked matrix in Dual Filter ($\alpha_{train} = 1.0, \alpha_{test} = 1.0$)}
    \label{fig:jaccard-1.0}
\end{figure*}

\begin{figure*}[ht!]
    \centering
    \includegraphics[width=0.65\linewidth]{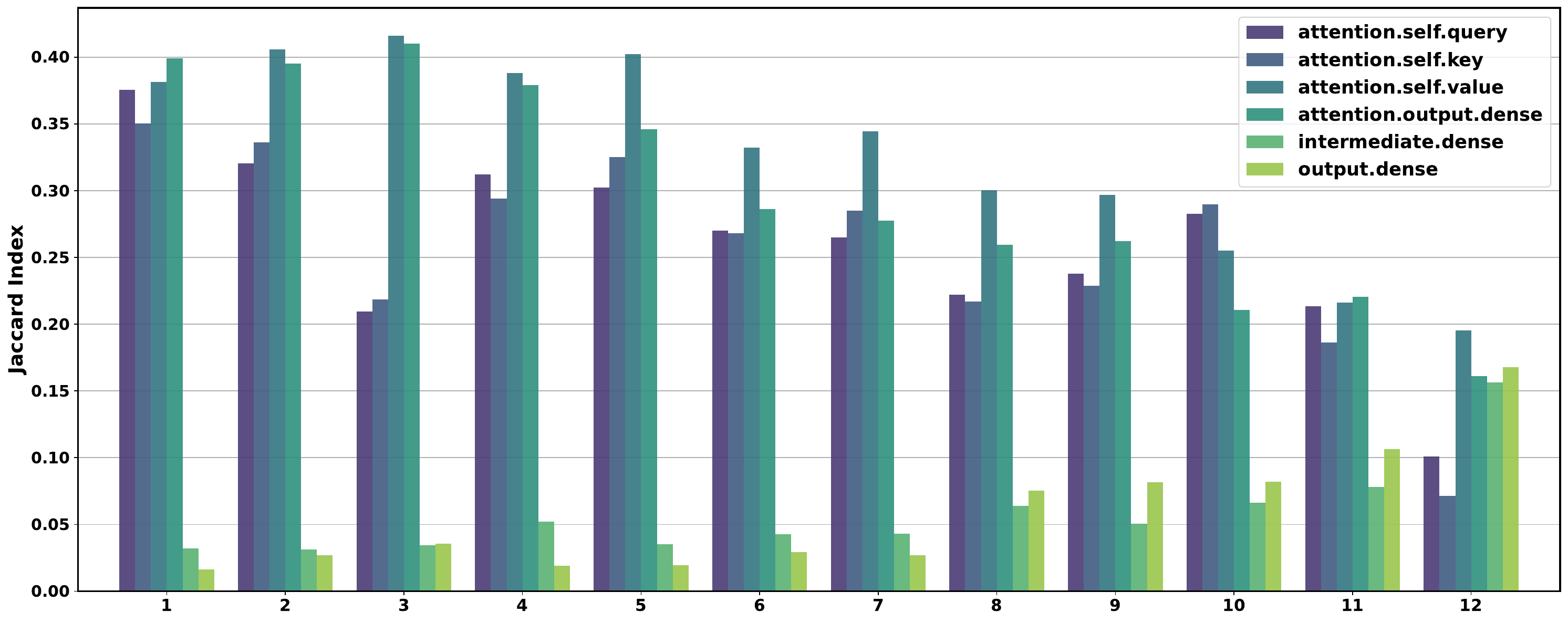}
    \caption{Jaccard Index for each of the tracked matrix in Dual Filter ($\alpha_{train} = 3.0, \alpha_{test} = 0.33$)}
    \label{fig:jaccard-3.0}
\end{figure*}

\begin{figure*}[ht!]
    \centering
    \includegraphics[width=0.65\linewidth]{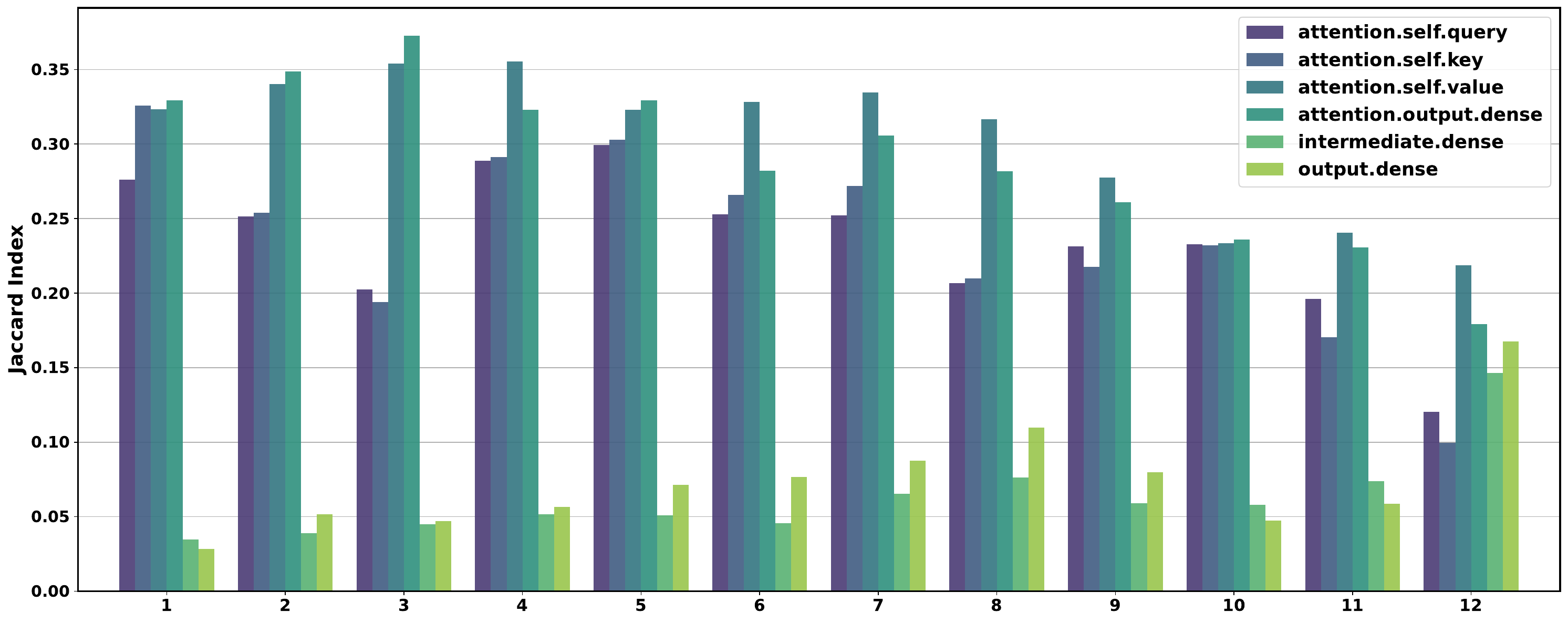}
    \caption{Jaccard Index for each of the tracked matrix in Dual Filter ($\alpha_{train} = 5.0, \alpha_{test} = 0.20$)}
    \label{fig:jaccard-5.0}
\end{figure*}

While the aim of this work is to eliminate gender confounding effects from the model's dementia detection capability, there is a possibility that the weights associated with dementia and gender become entangled during the learning process (i.e., same weights responsible for both - gender and dementia encoding). To investigate this, we record the change matrices for all layers in the network during the Dual Filter training process. We then conduct an analysis of the similarity between the change matrices from the fine-tuned dementia model and those from the fine-tuned gender model. For similarity measurements, we utilize the Jaccard Index to quantify the similarity between the two input matrices, which is defined as:
$$J(U,V) = \frac{|U \cap V|}{|U \cup V|}$$
To prepare the input, 85\% percentiles of the two change matrices are calculated and then the percentile values are used to binarize each of the matrices. Figure \ref{fig:jaccard-0.2} to ~\ref{fig:jaccard-5.0} demonstrates the barplot from six of the tracked weight matrices at each layer, with different $\alpha_{train}$  configurations. From the plots we can observe that at lower encoder layers, the similarity between dementia model and gender model concentrates on the attention block, especially $W_V$ and $W_O$. As we move up to the upper layer, the FFN block starts to display more similarity and jumps up at $12^{th}$ layer. Similar patterns are also observed in other $\alpha_{train}$ configurations. This result indicates the fine-tuned model stores information dynamically through the whole network and shift the storage at different layers. This finding also aligns with other work \citep{wei2024assessing} where weights entanglement are assessed with a larger model and different tasks.

\end{document}